\documentclass[11pt]{article}

\usepackage[final]{acl}
\usepackage{times}
\usepackage{latexsym}
\usepackage[T1]{fontenc}
\usepackage[utf8]{inputenc}
\usepackage{microtype}
\usepackage{inconsolata}
\usepackage{graphicx}
\usepackage{booktabs}
\usepackage{array}
\usepackage{amsmath}
\usepackage{amssymb}
\usepackage{url}
\usepackage{float}
\usepackage{dblfloatfix}
\usepackage{listings}
\usepackage{xcolor}

\definecolor{PromptBack}{HTML}{F6F6F6}

\lstnewenvironment{promptlisting}[1]{  \lstset{    title={\footnotesize\bfseries #1},
    basicstyle=\ttfamily\footnotesize,
    backgroundcolor=\color{PromptBack},
    frame=none,
    xleftmargin=0.5em,
    xrightmargin=0.5em,
    breaklines=true,
    breakatwhitespace=false,
    columns=fullflexible,
    keepspaces=true,
    showstringspaces=false,
    aboveskip=4pt,
    belowskip=4pt
  }}{}

\title{Goldsmith: Gold-Loss-Guided Definition Optimization with an Agentic Annotation Harness}

\author{
Yihan Li\textsuperscript{1,\ensuremath{\dagger},*} \quad
Hanyi Zhang\textsuperscript{2,\ensuremath{\dagger}} \quad
Xiaoxi Jiang\textsuperscript{1} \quad
Man Guo\textsuperscript{1,*}\\
\textsuperscript{1}Sun Yat-sen University, Guangzhou, China\\
\textsuperscript{2}South China University of Technology, Guangzhou, China\\
\texttt{liyihan.xyz@gmail.com}; \texttt{gman@mail.sysu.edu.cn}\\[2pt]
\textsuperscript{\ensuremath{\dagger}}Equal contribution. \textsuperscript{*}Corresponding authors.
}

\begin{document}
\maketitle
\begin{abstract}
Many annotation projects begin before experts have a stable guideline or enough labels to train a task-specific model.
We present Goldsmith, an agentic pipeline that turns a small gold set---expert-annotated calibration examples representing the intended task boundaries---into a reusable structured annotation definition.
Goldsmith treats this definition as a trainable textual object.
Candidate definitions are run on the same gold examples and scored with an executable structured loss, while the output schema, formatting, retrieval, repair, judging, and human review remain in an external harness.
A large language model (LLM) editor converts the highest-loss failures into textual-gradient revisions, which are accepted only when the measured loss decreases.
In prompt-optimization comparisons, Goldsmith improves over direct rewriting, OPRO, APE, and PromptBreeder under matched evaluation protocols.
The resulting definition also improves downstream annotation when combined with retrieval, score-based routing, and human review across typed span, pair-level relation, and fixed-trigger event-argument tasks.
These results show that scarce expert supervision can support both task-definition learning and scalable annotation.
\end{abstract}

\section{Introduction}

Many annotation projects begin before the task is fully defined: experts may recognize the phenomenon to annotate, yet lack a stable boundary policy, a complete manual, or enough labels for a task-specific model \citep{settles2009active,tomanek2009activeannotation}.
Pre-trained language model (PLM)-first systems remain strong once this target is stable.
BERT-style and XLM-R token classifiers are natural span-labeling baselines \citep{devlin2019bert,conneau2020xlmr}, while typed-marker relation classifiers can use stronger encoders such as RoBERTa or DeBERTa \citep{liu2019roberta,he2023debertav3}.
Even parameter-efficient updates such as Low-Rank Adaptation (LoRA) reduce training cost, not the cost of deciding what the task means \citep{hu2022lora}.
The bottleneck is therefore earlier: converting a partially articulated concept into an operational annotation procedure, a problem also discussed in LLM annotation and guideline-following IE \citep{tan2024annotationsurvey,sainz2024gollie}.

Long-context in-context learning changes this early-stage design space.
Few-shot prompting showed that examples can condition a model without weight updates \citep{brown2020language}; many-shot and long-context studies show that expert examples, failures, and revision history can now fit inside the working context \citep{agarwal2024manyshot,bertsch2024longcontexticl}.
However, unselected or unvalidated examples can dilute the expert signal \citep{zou2024manyshotlongcontexteval,zhang2025dricl}.
In this setting, carefully chosen gold examples can define the task boundary more concretely than an initial prose guideline.

\begin{figure}[t]
  \centering
  \includegraphics[width=\linewidth]{"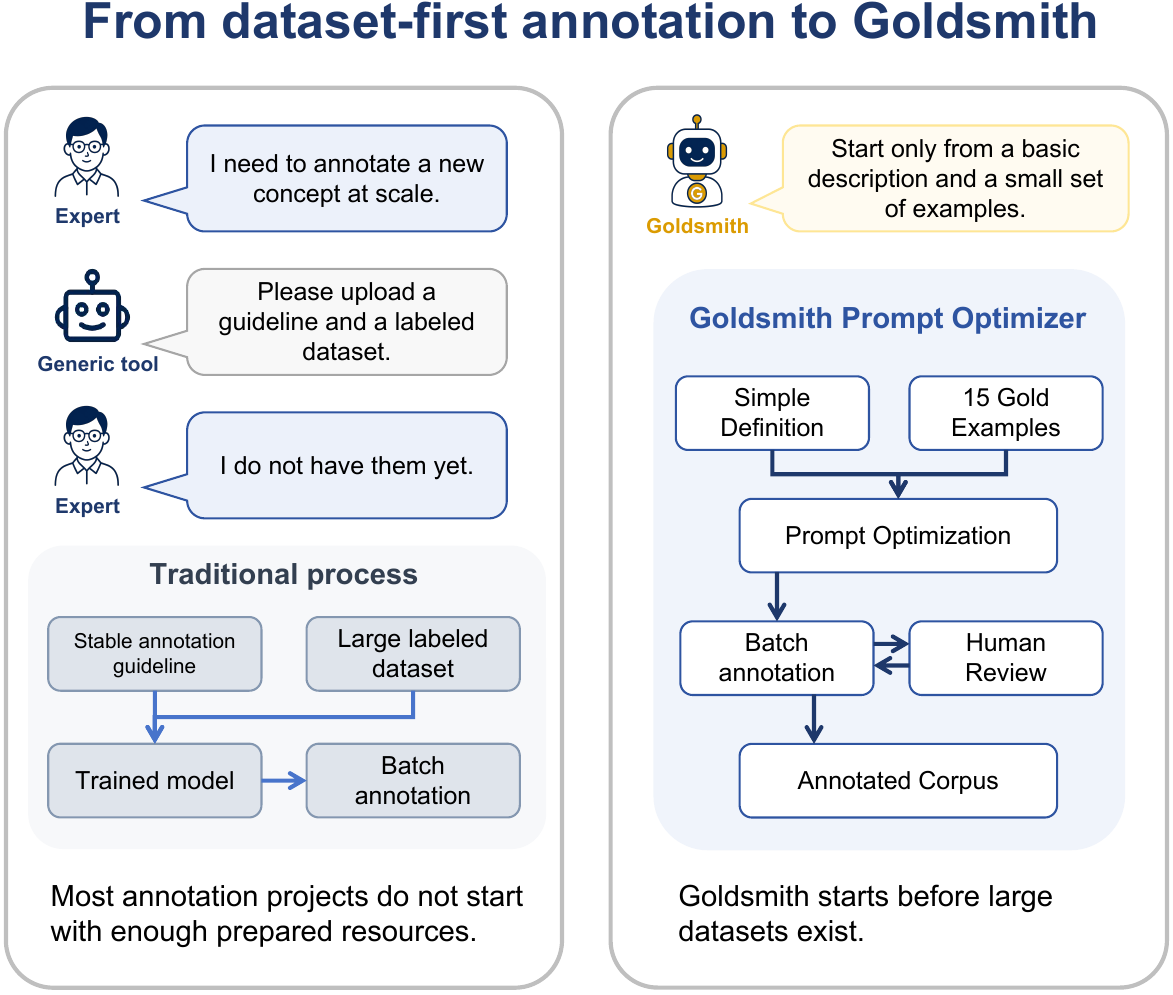"}
  \caption{Conceptual comparison between a dataset-first annotation workflow and Goldsmith's definition-first workflow.}
  \label{fig:teaser}
\end{figure}

We introduce Goldsmith, an annotation harness built around this premise.
Gold examples are the scarce material experts can provide early; the system repeatedly tests, reshapes, and polishes that material into a durable operational definition.
Given a short concept description and a 15-example gold calibration set, Goldsmith treats each gold example as a test case for the current definition, using prediction errors to decide whether a revised definition has captured the expert's intended boundary.
The definition then runs inside a model-external harness that connects model calls to retrieval tools, context construction, format repair, judges, review queues, and feedback loops; these tool-calling capabilities turn a revised definition into a stable workflow rather than a one-off LLM response \citep{liu2024toolace,qin2025metatool}.

The methodological core is prompt-as-parameter training: Goldsmith optimizes only the operational definition, while the schema, format reference, retrieved references, and harness controls remain outside the trainable definition.
Candidate definitions come from OPRO, LLM-derived textual gradients, or Goldsmith's own gold-loss-guided editor. This design is related to instruction-level optimizers such as APE and PromptBreeder, which generate or evolve candidate prompts \citep{zhou2022ape,fernando2023promptbreeder}, but acceptance is decided by the same gold loss rather than by the LLM's self-assessment \citep{yang2023opro,pryzant2023protegi,yuksekgonul2024textgrad}.
The claim is bounded: Goldsmith targets the early, evolving setting before PLM-first pipelines have enough stable labels, and asks whether the same 15 gold examples can outperform zero-shot prompting, static ICL, retrieval-only prompting, OPRO, and low-budget fine-tuning.

The contribution of this work can be summarized as follows:
\begin{itemize}
  \item We introduce a gold-loss-guided prompt optimization method that treats the operational definition as the trainable textual parameter and accepts LLM-proposed revisions only when they improve structured loss on expert gold examples.
  \item We build this optimizer into an annotation pipeline whose harness keeps schema, formatting, retrieval, repair, judge routing, and human review outside the trainable prompt while making the optimized definition usable for batch annotation.
  \item We empirically validate the approach across three structured annotation settings---OpenNER for typed multilingual named entity recognition (NER), CrossRE for pair-level relation classification, and PHEE V2 for controlled event-argument extraction---against LLM-only, prompt-optimization, and PLM baselines \citep{palen-michel-etal-2025-openner,bassignana2022crossre,phee2024chatgpt}.
\end{itemize}

\begin{figure*}[t]
  \centering
  \includegraphics[width=0.85\textwidth]{"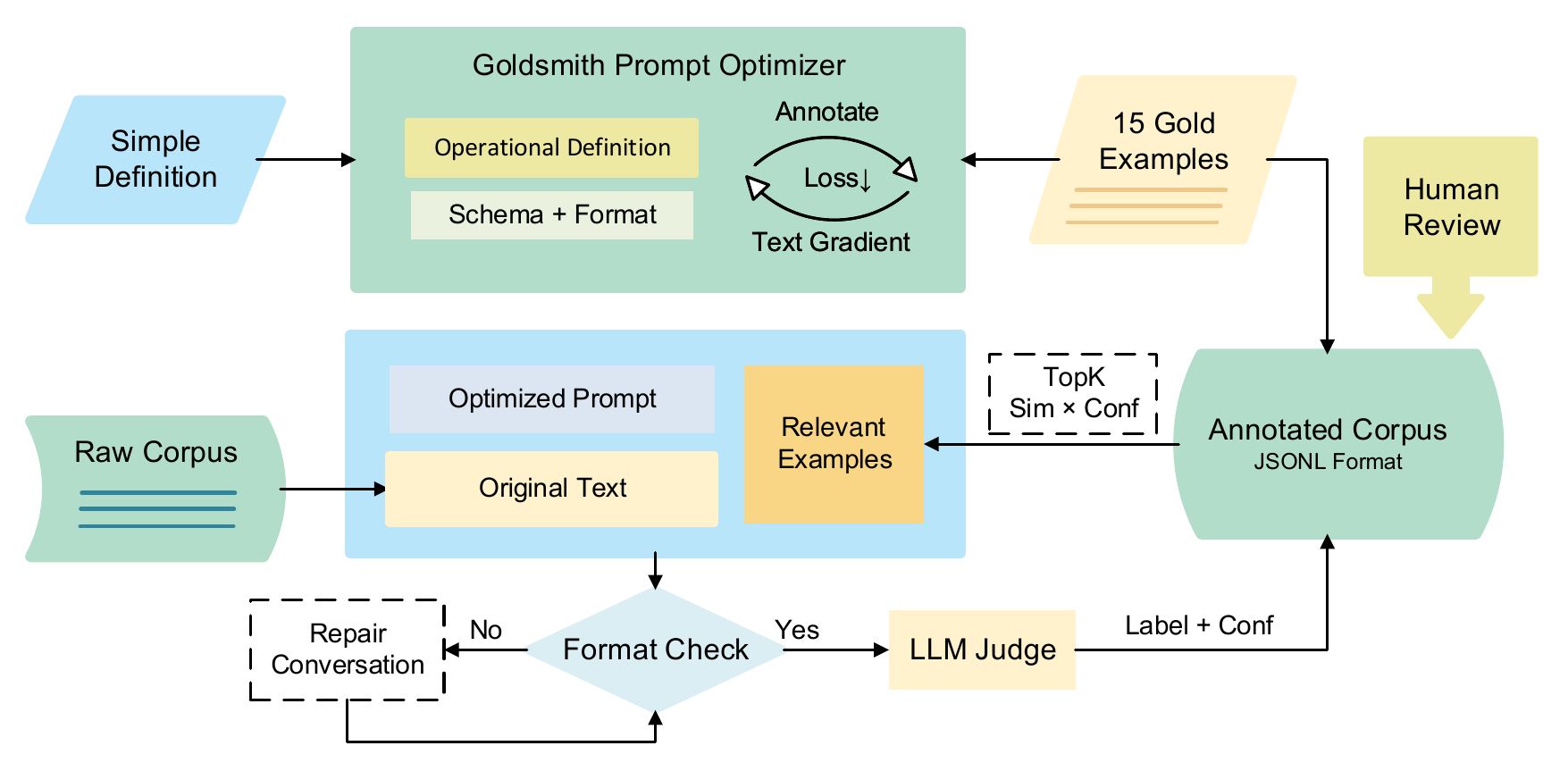"}
  \caption{Two-stage Goldsmith workflow: gold-loss-guided definition training followed by retrieval- and review-assisted batch annotation.}
  \label{fig:pipeline-overview}
\end{figure*}

\section{Related Work}

\subsection{LLM Annotation and Task Definition}
LLM annotation studies ask whether language models can replace or assist human labelers, while later active and analyst-in-the-loop systems use uncertainty and human checks to allocate expert effort \citep{wang2021want,gilardi2023chatgpt,zhang2023llmaaa,dai2023llmintheloop}.
Surveys and information-extraction work emphasize that generated labels need validation and that structured outputs remain sensitive to task guidelines, schemas, parsing, and repair \citep{tan2024annotationsurvey,xu2024generativeiesurvey,sainz2024gollie,huang2024clinical}.
Few-shot prompting and recent long-context studies make it possible to provide examples, failures, and revision context without updating model parameters \citep{brown2020language,agarwal2024manyshot,bertsch2024longcontexticl}.
However, more context is not automatically better supervision: demonstrations should be selected and validated, so examples serve as operational task-definition evidence rather than merely additional prompt tokens \citep{zou2024manyshotlongcontexteval,zhang2025dricl,zhang2025gradientmatchingicl}.

\subsection{Prompt Optimization}
Prompt optimization work moves beyond hand-written prompting; recent surveys describe a spectrum from discrete trigger search to LLM-driven instruction revision \citep{ramnath2025aposurvey}.
APE and PromptBreeder extend this idea to natural-language instructions: APE generates and scores candidate instructions, whereas PromptBreeder applies evolutionary mutation and self-referential prompt search \citep{zhou2022ape,fernando2023promptbreeder}.
OPRO frames an LLM as a black-box optimizer over prompt-score histories \citep{yang2023opro}.
ProTeGi uses a critic to convert failures into natural-language textual gradients and an editor to produce candidate prompts \citep{pryzant2023protegi}; TextGrad generalizes textual feedback as a differentiable-style signal for compound LLM systems \citep{yuksekgonul2024textgrad}.
These methods share an LLM proposal loop, but Goldsmith makes the operational definition the only trainable artifact and uses structured gold loss under a fixed annotation interface as the candidate acceptance rule.

\section{Method}
\label{sec:method}

Goldsmith turns a small set of expert gold examples into a reusable structured annotation definition.
The system has two stages: it first trains and selects an operational definition on a fixed gold set, then applies the resulting definition to an unlabeled corpus through retrieval, format checking, quality routing, and human review.
Given an initial task description \(c_0\), a gold set \(G\), and an unlabeled corpus \(U\), Goldsmith produces an optimized definition \(c^\star\) and structured predictions with confidence scores on \(U\).
Only the operational definition is optimized during prompt training; the label inventory, output schema, parser, format repair, retrieval, and review policies are managed by the external harness.
Figure~\ref{fig:pipeline-overview} shows the complete workflow.

\subsection{Gold Loss for Structured Annotation}

For any candidate definition \(c\), Goldsmith combines it with the fixed output protocol to form an annotation prompt and runs that prompt on the same gold set.
Model outputs are first parsed and checked against the structural constraints, then compared item by item with expert gold annotations.
The evaluator computes a structured error for each item and averages these errors into the candidate's gold loss.
Thus, structured gold loss is an executable objective computed from fixed gold annotations and comparison rules, rather than a subjective score supplied by an LLM or judge.

For typed span annotation, each item contains a set of labeled spans.
Let \(G_i^T\) and \(P_i^T\) denote gold and predicted typed spans \((s,e,y)\), and let \(G_i^B,P_i^B\) denote their boundary-only projections.
We define normalized missed- and extra-boundary rates as \(\mathrm{FN}_B(i)\) and \(\mathrm{FP}_B(i)\), respectively.
Boundary loss combines these rates with task-level weights:
\[
L_B(i)=
\frac{\alpha_{\mathrm{miss}}\,\mathrm{FN}_B(i)
      +\alpha_{\mathrm{extra}}\,\mathrm{FP}_B(i)}
     {\alpha_{\mathrm{miss}}+\alpha_{\mathrm{extra}}}.
\]
Here \(\mathrm{FN}_B(i)=|G_i^B\setminus P_i^B|/\max(1,|G_i^B|)\) and \(\mathrm{FP}_B(i)=|P_i^B\setminus G_i^B|/\max(1,|P_i^B|)\). The weights allow the evaluation to place different costs on missed and extra boundaries.
Type loss is computed only on matched boundaries, \(M_i=G_i^B\cap P_i^B\):
\[
L_{\mathrm{type}}(i)=
\begin{cases}
E_{\mathrm{type}}(i)/|M_i|, & |M_i|>0,\\
0, & |M_i|=0.
\end{cases}
\]
We combine these components as
\[
L_{\mathrm{span}}(i)=\lambda_B L_B(i)+\lambda_T L_{\mathrm{type}}(i),
\]
The normalized weights satisfy \(\lambda_B+\lambda_T=1\) and express the relative importance of boundary recovery and type correctness; additional item-level penalties are described below.

For given-pair relation annotation, the input supplies a sentence and a marked head--tail entity pair, and the model selects one relation label, including \texttt{no\_relation}.
The per-item relation loss is label disagreement:
\[
\tau_{\mathrm{rel}}(i)=\mathbb{1}[\hat{y}_i\neq y_i].
\]
For relation extraction with multiple relation edges, we analogously compare normalized missed- and extra-relation rates using \(\alpha_{\mathrm{miss}}\) and \(\alpha_{\mathrm{extra}}\). Parse failure and instability are handled as item-level penalties below.
We denote the resulting relation loss by \(L_{\mathrm{rel}}(i)\); for a single given pair, \(L_{\mathrm{rel}}(i)=\tau_{\mathrm{rel}}(i)\).
For an interface that combines spans and relations, such as the fixed-trigger PHEE V2 setting, Goldsmith combines the corresponding span and relation losses and checks the trigger-to-argument contract.
We write this task loss as
\[
L_{\mathrm{task}}(i)=
\begin{cases}
L_{\mathrm{span}}(i), & \text{span},\\
L_{\mathrm{rel}}(i), & \text{relation},\\
\lambda_S L_{\mathrm{span}}(i)+\lambda_R L_{\mathrm{rel}}(i),
  & \text{joint}.
\end{cases}
\]
Here, span, relation, and joint denote span-only, relation-only, and span-relation interfaces, respectively, and \(\lambda_S+\lambda_R=1\).

The final item-level error is defined piecewise so that parsing failures are part of the same objective:
\[
\ell_i(c)=
\begin{cases}
\lambda_{\mathrm{parse}},
  & \text{parse failure},\\
L_{\mathrm{task}}(i)+P_i(c),
  & \text{otherwise}.
\end{cases}
\]
where the applicable non-format penalties are collected as
\[
\begin{aligned}
P_i(c)={}&\lambda_{\mathrm{exact}}\mathbb{1}[\text{exact mismatch}]\\
&+\lambda_{\mathrm{stab}}\mathbb{1}[\text{unstable output}]\\
&+\lambda_{\mathrm{trigger}}\mathbb{1}[\text{trigger failure}].
\end{aligned}
\]
Each indicator and penalty is included only when the corresponding contract applies. For relation-only classification, label disagreement is already the task loss and is not counted again as a separate exact-match penalty.
The prompt-level objective is
\[
\bar L(G,c)=\frac{1}{|G|}\sum_{i=1}^{|G|}\ell_i(c).
\]
For reporting, the implementation may use the monotonic rescaling \(L(G,c)=\rho\,\bar L(G,c)\), where \(\rho\) is a reporting-scale parameter. All candidate definitions use the same objective for comparison. Judge scores, retrieval similarity, and human review outcomes do not replace this training objective.

\subsection{Gold-Loss-Guided Definition Training}
\begin{figure}[h]
  \centering
  \includegraphics[width=\linewidth]{"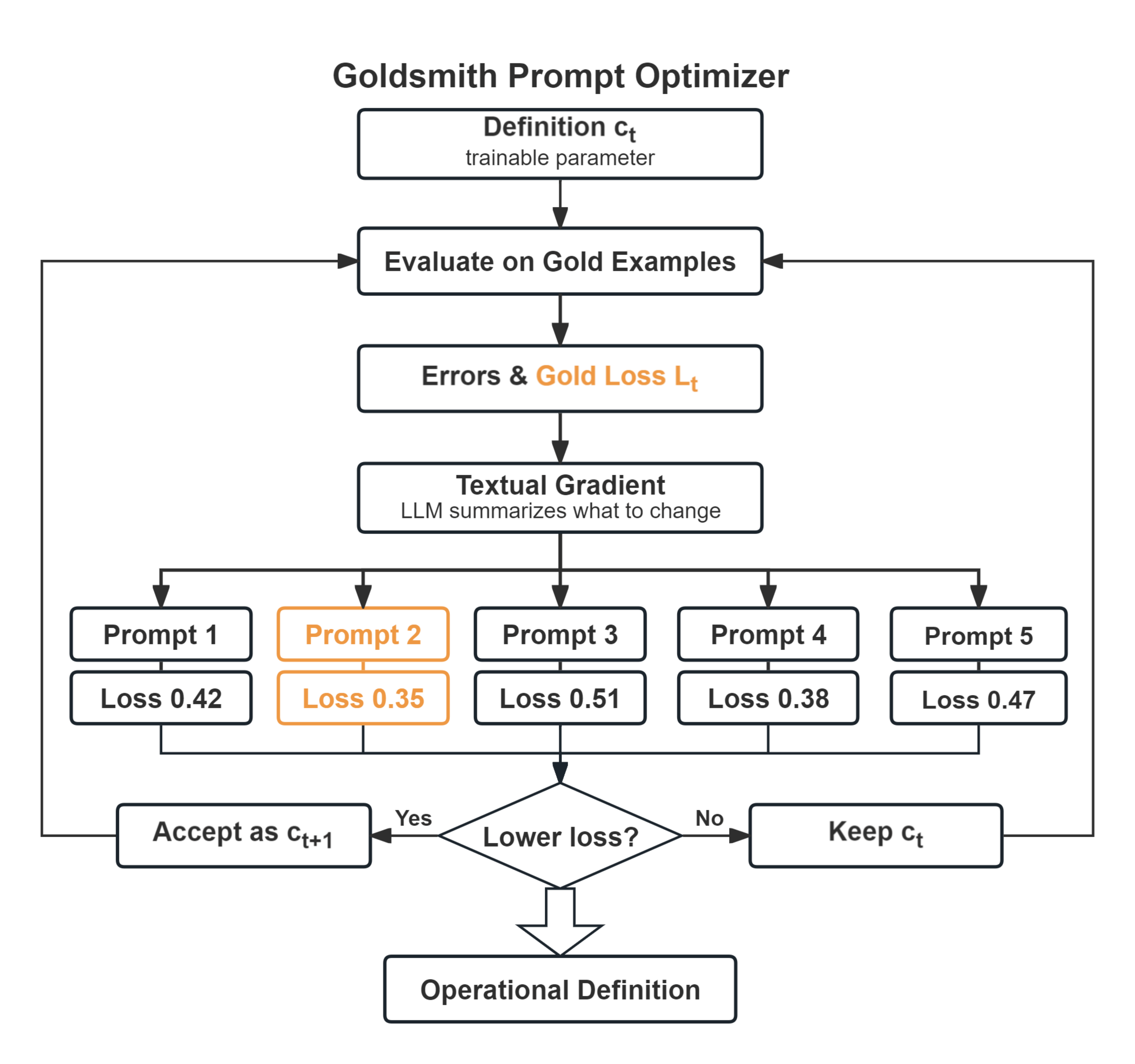"}
  \caption{Goldsmith definition-training loop: textual-gradient edits are evaluated on gold examples and accepted only when they reduce gold loss.}
  \label{fig:prompt-optimizer}
\end{figure}

Let \(c_t\) denote the operational definition at optimization round \(t\), and let \(L_t=L(G,c_t)\) denote its gold loss under the objective in Section~\ref{sec:method}.
Given an input \(x\), retrieved references \(r\), schema \(q\), and format reference \(f\), the annotation prompt is \(p_t=h(c_t,x,r,q,f)\).
Goldsmith optimizes only \(c_t\), which specifies what should be annotated, how boundaries should be determined, and how labels or relations should be interpreted.
The schema, format reference, parser, repair policy, retrieval policy, and review policy remain fixed.
Each candidate is therefore evaluated through the same annotation interface and against the same structured gold-loss objective.

Goldsmith builds on the textual-gradient feedback of ProTeGi and the textual optimization view of TextGrad \citep{pryzant2023protegi,yuksekgonul2024textgrad}, but its contribution is not textual gradients themselves.
Instead, Goldsmith uses them to optimize a structured annotation definition from a small expert gold set.
The optimization target is the operational definition rather than an unconstrained prompt, and candidate definitions must express abstract boundary rules, label priorities, and decision criteria rather than copy instance-specific examples.
At each round, Goldsmith evaluates the current definition \(c_t\) on all gold items, selects the \(k\) highest-loss failures, and summarizes them into edit directions.
An LLM editor proposes candidate definitions, but does not determine whether they are better: each candidate is rerun on \(G\) through the same parser and schema, and is accepted only when its measured structured gold loss decreases.

The optimization process is stateful.
Goldsmith records accepted and rejected candidates, their measured losses, the highest-loss failures before each edit, and the residual failures after accepted edits.
This optimization trace provides the editor with evidence about which semantic changes improve the shared objective and which hard cases remain unresolved, reducing repeated ineffective rewrites.
Training stops when all gold items pass or when the patience budget is exhausted.

\subsection{Batch Annotation and Review Harness}

After training, Goldsmith applies \(c^\star\) to \(U\) through the fixed model-external harness.
For each item, the context builder retrieves eligible references from the gold set and previously reviewed annotations.
The annotator then produces a structured prediction using \(c^\star\), the retrieved context, and the fixed output protocol.

The parser checks schema fields, label inventory, text preservation, span validity, relation endpoints, and other format constraints.
If a response violates only the format contract, Goldsmith first performs conversational repair that asks the model to preserve the annotation content while correcting the structure.
Only unrepaired failures are retried as full annotation attempts.

An LLM judge can score parseable predictions for format validity, concept fit, boundary or relation quality, missed-span risk, extra-span risk, and overall quality.
Judge scores are used only as review-priority signals.
Before relying on them for routing, Goldsmith checks that low-score bins concentrate true errors.

Human review prioritizes low-score or otherwise high-risk items.
In benchmark runs, selected items are corrected using the available gold annotations to simulate reviewed feedback; in deployment, this step must be performed by domain experts.
Reviewed annotations can be promoted to reviewed references for later context construction and second-pass annotation, while definition updates remain subject to the independent gold-loss training procedure.

\section{Experiments}
\subsection{Tasks and Datasets}
\begin{table*}[t]
  \centering
  \footnotesize
  \setlength{\tabcolsep}{3pt}
  \begin{tabular*}{\textwidth}{@{\extracolsep{\fill}}p{0.14\textwidth}p{0.20\textwidth}p{0.24\textwidth}p{0.30\textwidth}@{}}
    \toprule
    Dataset & Task & Input & Output \\
    \midrule
    OpenNER
      & Typed spans
      & Text
      & PER / ORG / LOC \\
    CrossRE
      & Relations
      & Text + HEAD/TAIL pair
      & Relation label \\
    PHEE V2
      & Event arguments
      & Text + trigger
      & Subject / Treatment / Effect \\
    \bottomrule
  \end{tabular*}
  \caption{Task interfaces used in the OpenNER, CrossRE, and PHEE V2 experiments.}
  \label{tab:dataset-overview}
\end{table*}

We evaluate Goldsmith on three structured annotation settings.
OpenNER is a multilingual typed span-labeling task: the input is a text and the output is a set of entity spans labeled as \(\mathrm{PER}\), \(\mathrm{ORG}\), or \(\mathrm{LOC}\) \citep{palen-michel-etal-2025-openner}.
It represents span-labeling tasks because the system must recover both entity boundaries and type labels.
CrossRE is a given-pair relation-classification task: the input is a sentence with a marked head entity, tail entity, and entity types, and the output is one of 17 positive relation labels or \texttt{no\_relation} \citep{bassignana2022crossre}.
It represents pair-level relation classification.
PHEE V2 provides a controlled event-argument setting associated with pharmacovigilance event extraction \citep{phee2024chatgpt}.
The event trigger is supplied as a fixed anchor, and the model predicts \texttt{Subject}, \texttt{Treatment}, and \texttt{Effect} argument spans together with trigger-to-argument relations.
This setting tests argument-role boundary decisions without requiring the model to discover event triggers.

For each setting, we run experiments on a fixed dataset slice and reserve a 15-example calibration set for prompt optimization where applicable.
The OpenNER multilingual slice and gold selection are described in Appendix~\ref{app:openner-slice} and Appendix~\ref{app:openner-gold}; the CrossRE six-domain slice and pair-level gold selection are described in Appendix~\ref{app:crossre-slice} and Appendix~\ref{app:crossre-gold}.
Unless otherwise specified, all LLM-based prompt optimization, annotation, judging, and review-routing experiments use GPT-5.5; model identifiers, access month, decoding parameters, retry bounds, and review settings are listed in Appendix~\ref{app:model-runtime}.
For OpenNER task-stage evaluation, all rows use the same held-out evaluation subset.
The 15 calibration examples and the items selected for human review are excluded from this subset for every setting.
In review-based settings, selected low-score items are corrected as reviewed feedback, promoted to reviewed references, and then used when the system reruns annotation over the target corpus.

\subsection{Prompt-Optimization Baselines}
The prompt-optimization experiment tests whether textual-gradient feedback improves the operational definition from 15 gold examples.
All optimizers share the same initial description, gold set, candidate budget, gold loss, acceptance rule, and patience rule.

\paragraph{LLM rewrite-only.}
This baseline receives only the current operational definition and a generic rewrite instruction, without gold examples, errors, history, or textual-gradient feedback.

\paragraph{OPRO.}
OPRO is implemented as a history-based black-box optimizer \citep{yang2023opro}.
It sees previous candidate definitions and loss-derived scores, but not source texts, gold answers, model outputs, or error types.
Candidates are accepted only if measured gold loss improves.

\paragraph{APE.}
APE uses its candidate-generation and scoring procedure, with the same GPT-5.5 backend, 15-example calibration set, parser/schema, and final structured gold-loss objective as the other optimizers \citep{zhou2022ape}.

\paragraph{PromptBreeder.}
PromptBreeder is a budget-matched reproduction of the evolutionary search procedure, evaluated under the same backend, calibration set, parser/schema, and gold-loss objective \citep{fernando2023promptbreeder}.

\paragraph{Goldsmith.}
Goldsmith is the method described in Section~\ref{sec:method}.
It summarizes the highest-loss gold errors into textual-gradient edit directions, maintains optimization memory across rounds, asks an LLM editor for abstract candidate definitions, and accepts a candidate only when the same gold loss decreases.
We report loss, 15-gold pass count, best round, evaluated candidates, elapsed time, and token usage.

\subsection{Task-Stage Baselines}
Task-stage baselines compare the resulting annotation workflow with low-budget supervised PLM baselines under the same task interface where possible.
For OpenNER, Goldsmith score-routed review uses the optimized definition followed by score-routed review, while Goldsmith without review reports the first annotation pass before human correction.
Goldsmith zero-shot uses the optimized definition but disables retrieval, judge scoring, and review, isolating the value of the learned definition alone.
Initial-prompt zero-shot uses 0 gold examples and the initial task description before prompt optimization, also without retrieval, judge scoring, or review.
The PLM reference condition follows the local gold4-per-language setup: mBERT and XLM-R token classifiers are fine-tuned on 208 deterministic multilingual training examples.
For CrossRE, Zero-shot LLM receives the sentence, marked head--tail pair, entity types, and relation-label set, but no gold examples or optimized definition.
The PLM comparison uses typed-marker relation classifiers: the official BERT CrossRE anchor and stronger RoBERTa-large or DeBERTa-v3 encoder variants, all receiving the same sentence, gold entity pair, entity types, and relation-label set \citep{bassignana2022crossre,devlin2019bert,liu2019roberta,he2023debertav3}.
For CrossRE, all rows are evaluated on the same 29,171-pair evaluation slice.
Goldsmith uses the optimized relation definition with reviewed feedback and retrieved references, while the PLM rows are low-budget supervised references trained from 200 sentences expanded into 10,593 training pairs.
For PHEE V2, we compare initial zero-shot prompting with Goldsmith before and after score-routed review using argument micro-F1, relation micro-F1, trigger pass rate, and joint exact match.

\subsection{Metrics}
Prompt-optimization results report gold-set pass count, gold loss, best round, and evaluated candidate count.
OpenNER task-stage results report supervision or feedback used by the system, typed precision, typed recall, micro-F1, macro-F1, and typed exact match on the shared held-out evaluation subset.
CrossRE reports accuracy, macro-F1 over the positive relation classes, positive-relation recall, and positive-relation F1; the macro-F1 calculation excludes the dominant negative class.
PHEE V2 reports argument and relation micro-F1, trigger pass rate, and joint exact match.
OpenNER component ablations report precision, recall, F1, and F1 change relative to the main setting, while model ablations report F1 without review, F1 with review, and exact match.

\begin{table*}[h]
  \centering
  \footnotesize
  \setlength{\tabcolsep}{3pt}
  \begin{tabular*}{\textwidth}{@{\extracolsep{\fill}}llccccc@{}}
    \toprule
    Setting & Supervision / feedback & P & R & F1 & Macro & Exact \\
    \midrule
    Goldsmith score-routed review & 15 gold + 193 review & \textbf{0.8092} & \textbf{0.8257} & \textbf{0.8174} & \textbf{0.8919} & \textbf{0.8140} \\
    Goldsmith without review & 15 gold & 0.7293 & 0.7505 & 0.7398 & 0.8669 & 0.7749 \\
    Goldsmith zero-shot & 15 gold & 0.7133 & 0.7440 & 0.7283 & 0.8556 & 0.7613 \\
    Initial-prompt zero-shot & 0 gold & 0.7040 & 0.7441 & 0.7235 & 0.8546 & 0.7565 \\
    mBERT & 208 ex. & 0.4867 & 0.5595 & 0.5206 & 0.5413 & 0.5630 \\
    XLM-R & 208 ex. & 0.3281 & 0.3799 & 0.3521 & 0.3735 & 0.4210 \\
    \bottomrule
  \end{tabular*}
  \caption{OpenNER task-stage results on a shared held-out subset, comparing Goldsmith review settings, zero-shot prompting, and low-budget PLM baselines.}
  \label{tab:openner-main-results}
\end{table*}

\section{Results and Analysis}

\subsection{Prompt Optimization Results}

Table~\ref{tab:prompt-training-results} isolates the definition-training problem before any held-out task evaluation.
All optimizers start from the same short description, the same 15 gold examples, and the same candidate budget.
Direct LLM rewriting improves the prompt but stalls at \(7/15\) passing gold items; OPRO~\citep{yang2023opro} does better at \(9/15\), while APE~\citep{zhou2022ape} and PromptBreeder~\citep{fernando2023promptbreeder} reach \(10/15\) and \(12/15\), respectively.
Goldsmith reaches \(14/15\) with loss \(1.8030\), a much larger improvement under the same budget.
The main difference is not that Goldsmith uses a stronger editor, but that it converts the highest-loss gold failures into textual-gradient edit directions and still accepts edits only after measured gold-loss reduction.
This result establishes optimization efficiency on the shared gold-loss objective; the downstream ablation in Table~\ref{tab:openner-component-ablation} then checks whether the optimized definition remains useful inside the annotation workflow.

\begin{table}[H]
  \centering
  \footnotesize
  \setlength{\tabcolsep}{4pt}
  \begin{tabular*}{\columnwidth}{@{\extracolsep{\fill}}lrrr@{}}
    \toprule
    Optimizer & Pass & Loss $\downarrow$ & Best rnd./gen. \\
    \midrule
    LLM rewrite-only & 7/15 & 22.0303 & 7 \\
    OPRO & 9/15 & 14.5909 & 16 \\
    APE & 10/15 & 11.5606 & N/A \\
    PromptBreeder & 12/15 & 8.9091 & 23 gen. \\
    Goldsmith & \textbf{14/15} & \textbf{1.8030} & 18 \\
    \bottomrule
  \end{tabular*}
  \caption{OpenNER prompt-optimization results under a shared initial definition and 15-example gold calibration set. Pass denotes the number of gold examples solved, and lower gold loss is better.}
  \label{tab:prompt-training-results}
\end{table}

\begin{figure}[h]
  \centering
  \includegraphics[width=\linewidth]{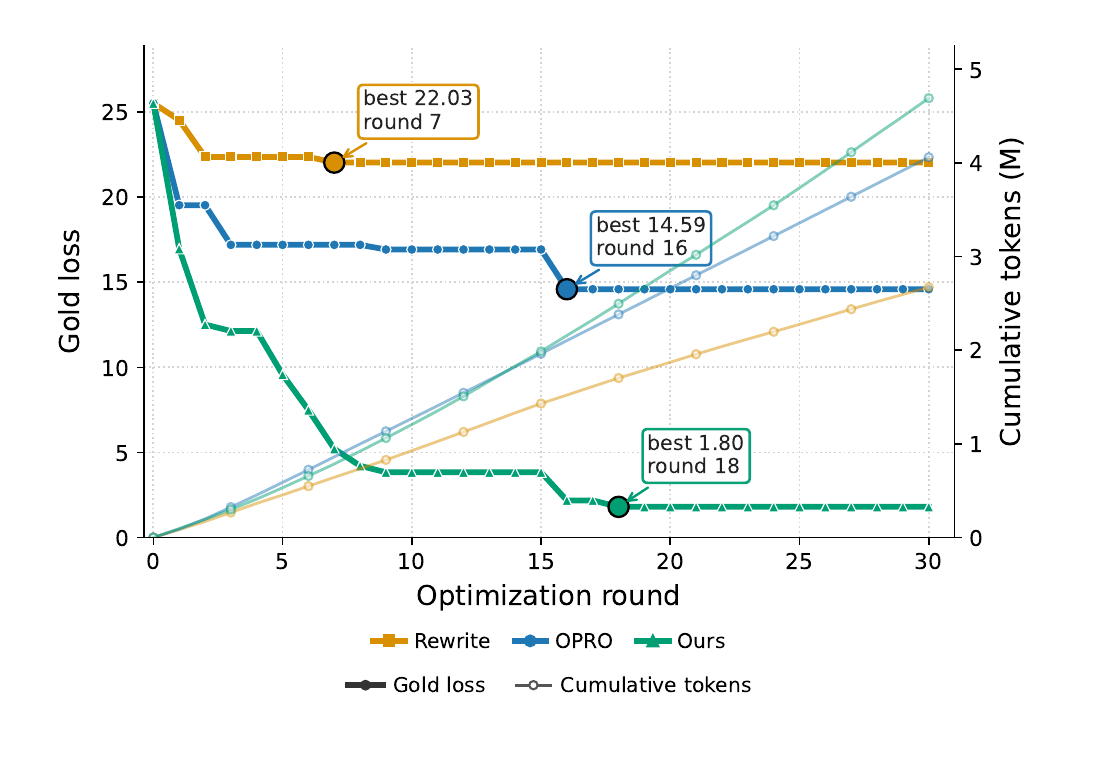}
  \caption{OpenNER gold-loss and token-use trajectories for LLM rewrite-only, OPRO, and Goldsmith. Thick lines show gold loss and thin lines show cumulative token usage over optimization rounds.}
  \label{fig:prompt-loss-curve}
\end{figure}

Figure~\ref{fig:prompt-loss-curve} shows the same pattern over time.
The baselines occasionally find better candidates, but their trajectories remain noisy because they receive either no diagnostic signal or only aggregate history.
Goldsmith produces a steadier loss drop, which is the behavior needed in the definition-building stage: the optimizer must turn a few expert examples into a more precise operational definition rather than merely generate fluent prompt variants.
\subsection{Task Results}

\paragraph{OpenNER.}
Table~\ref{tab:openner-main-results} reports the multilingual typed span-labeling result.
All rows use the same held-out evaluation subset, after excluding the 15 calibration examples and the reviewed feedback items.
The without review row evaluates the optimized definition after the first annotation pass.
The score-routed review row first corrects low-judge-score items as reviewed feedback, promotes them to reviewed references, and reruns annotation over the target corpus before evaluating on the same subset.
Under this second-pass protocol, Goldsmith reaches \(0.8174\) micro-F1, \(0.8919\) macro-F1, and \(0.8140\) typed exact match.
This supports the intended division of labor.
Prompt training makes the definition usable, while judge routing and review concentrate expert effort on likely errors instead of treating all items equally.

The same table also shows why the comparison is not simply ``LLM versus PLM.''
The optimized-definition zero-shot row improves over the initial-prompt zero-shot row, but both trail the normal without review run with retrieved references.
This separates two effects: prompt optimization improves the operational definition, while retrieval still provides useful in-context task grounding.
The low-budget mBERT and XLM-R rows are weaker under this data budget, which is consistent with the paper's target setting: supervised encoders become attractive after the task is stable and more labels exist, but they are a poor fit for the earlier definition-building phase.
These PLM numbers should therefore be read as budget-matched baselines, not as reproductions of full-data results reported in prior supervised NER work; their training data are deliberately limited to the same order of magnitude as the gold and review budget used by the annotation workflow.

\begin{table}[h]
  \centering
  \scriptsize
  \setlength{\tabcolsep}{4pt}
  \begin{tabular*}{\columnwidth}{@{\extracolsep{\fill}}lcccc@{}}
    \toprule
    Setting & Acc. & Pos. Macro & Pos. R & Pos. F1 \\
    \midrule
    Goldsmith & 0.8650 & \textbf{0.4273} & \textbf{0.7273} & \textbf{0.5439} \\
    Zero-shot LLM & 0.8720 & 0.1640 & 0.1980 & 0.2470 \\
    BERT & 0.3402 & 0.0883 & 0.4845 & 0.1307 \\
    RoBERTa & 0.3998 & 0.1108 & 0.5231 & 0.1511 \\
    DeBERTa & 0.4174 & 0.1088 & 0.4834 & 0.1445 \\
    \bottomrule
  \end{tabular*}
  \caption{CrossRE pair-level relation-classification results on the shared 29,171-pair evaluation slice.}
  \label{tab:plm-budget-curve}
\end{table}

\paragraph{CrossRE.}
Table~\ref{tab:plm-budget-curve} reports pair-level relation classification on the same 29,171-pair CrossRE evaluation slice.
This removes the denominator ambiguity: Goldsmith and the PLM anchors are compared on the same final evaluation scope, although they use different learning mechanisms.
Accuracy is not the decisive metric here because \texttt{no\_relation} dominates the test pairs.
Goldsmith reaches accuracy \(0.8650\), positive macro-F1 \(0.4273\), positive recall \(0.7273\), and positive F1 \(0.5439\).
The result suggests that the pipeline is not limited to span boundaries.
When the interface changes from span recovery to relation classification, the same principle still matters: a small set of reviewed examples and prompt-training feedback can steer the model toward rare positive labels that accuracy-oriented baselines tend to miss.

\paragraph{PHEE V2.}
PHEE V2 extends the evaluation to controlled event-argument extraction with a fixed trigger anchor.
The initial zero-shot setting reaches argument micro-F1 \(0.6637\), relation micro-F1 \(0.6620\), trigger pass rate \(0.9988\), and joint exact match \(0.3513\).
Goldsmith without review improves these values to \(0.7111\), \(0.7066\), \(1.0000\), and \(0.3958\), respectively.
Score-routed review further reaches argument micro-F1 \(0.7471\), relation micro-F1 \(0.7427\), trigger pass rate \(0.9988\), and joint exact match \(0.4500\).
This result supports cross-task applicability across span, pair-relation, and fixed-trigger argument interfaces.

\subsection{Ablation Experiments}

\begin{table}[h]
  \centering
  \footnotesize
  \setlength{\tabcolsep}{3pt}
  \begin{tabular*}{\columnwidth}{@{\extracolsep{\fill}}lrrrr@{}}
    \toprule
    Setting & P & R & F1 & \(\Delta\)F1 \\
    \midrule
    Ours & 0.8092 & 0.8257 & 0.8174 & -- \\
    Initial def. & 0.6510 & 0.6820 & 0.6661 & -0.1513 \\
    Random-10 ref. & 0.7832 & 0.8001 & 0.7915 & -0.0259 \\
    \bottomrule
  \end{tabular*}
  \caption{OpenNER ablations of definition optimization and reference selection, reported with precision, recall, and micro-F1.}
  \label{tab:openner-component-ablation}
\end{table}

Table~\ref{tab:openner-component-ablation} separates definition quality from reference selection.
All ablation rows use the same OpenNER evaluation subset as Table~\ref{tab:openner-main-results}.
Removing prompt optimization drops F1 from \(0.8174\) to \(0.6661\), indicating that the learned operational definition is not replaceable by the initial prose description.
Together with the calibration loss result in Table~\ref{tab:prompt-training-results}, this shows both that Goldsmith finds lower-loss definitions more efficiently and that using the optimized definition matters for the downstream task-stage run.
Replacing structured retrieval with random references is less damaging but still reduces F1 to \(0.7915\).
The two ablations therefore support different parts of the method: gold-loss prompt training supplies the main task-boundary improvement, while retrieval quality supplies additional stability during batch annotation.

\begin{table}[h]
  \centering
  \scriptsize
  \setlength{\tabcolsep}{2pt}
  \begin{tabular*}{\columnwidth}{@{\extracolsep{\fill}}llrrr@{}}
    \toprule
    Backend & Provider & \begin{tabular}{@{}c@{}}w/o\\review\end{tabular} & \begin{tabular}{@{}c@{}}w/\\review\end{tabular} & Exact \\
    \midrule
    GPT-5.5 & OpenAI & 0.7398 & \textbf{0.8174} & \textbf{0.8140} \\
    GPT-5.4 & OpenAI & 0.7222 & 0.8012 & 0.7989 \\
    Kimi K2.5 & Moonshot & 0.7160 & 0.7890 & 0.7820 \\
    DSv4p & DeepSeek & \textbf{0.7660} & 0.7947 & 0.7863 \\
    \bottomrule
  \end{tabular*}
  \caption{OpenNER backend ablation using the same optimized definition, with and without score-routed review.}
  \label{tab:openner-model-ablation}
\end{table}

Table~\ref{tab:openner-model-ablation} asks whether the result is tied to one backend.
GPT-5.5 and GPT-5.4 are OpenAI models, Kimi K2.5 is from Moonshot AI, and DSv4p is DeepSeek V4 Pro from DeepSeek \citep{openai2026gpt55,openai2026gpt54,kimi2026k25,deepseek2026models}; the exact run-facing identifiers and access month are listed in Appendix~\ref{app:model-runtime}.
The comparison keeps the optimized definition, review protocol, and held-out evaluation setting fixed across backends.
The reviewed ranking broadly follows external model-capability leaderboards such as the Artificial Analysis Intelligence Index, which aggregates reasoning, coding, mathematics, and knowledge benchmarks \citep{artificialanalysis2026}.
Model quality still matters: GPT-5.5 gives the best reviewed F1 and exact match, while DSv4p has the strongest without review F1 but loses that lead after review.
At the same time, the reviewed scores are relatively close across capable backends.
This is the desired behavior of the harness: optimized definitions, retrieval, repair, judge routing, and human review do not eliminate model differences, but they reduce the extent to which the final annotation quality is determined by raw backend strength alone.

\section{Conclusion}

Goldsmith addresses a narrow but common stage of annotation: experts know the phenomenon they want to annotate, but the operational definition is still being formed and only a few authoritative examples are available.
The central result is that those examples can be used as executable supervision rather than only as static demonstrations.
In prompt optimization, gold-loss-guided textual-gradient editing produces a substantially stronger operational definition than direct LLM rewriting or OPRO.
In task-stage evaluation, the resulting definition, combined with retrieval, judge routing, and human review, improves OpenNER typed span labeling, CrossRE positive-relation classification, and PHEE V2 fixed-trigger event-argument extraction under low-label budgets.

The broader implication is that long-context LLM annotation should not be reduced to adding more examples to a prompt.
Its practical value depends on a harness that keeps the schema fixed, tests definition edits against gold loss, repairs and audits model outputs, and sends uncertain cases back to human review.
This does not replace supervised PLM pipelines once large stable datasets exist.
Instead, it provides a way to reach that stable task definition more efficiently, with explicit traces of which gold examples shaped the definition and where human review changed the corpus.

\section*{Limitations}

Goldsmith is designed for tasks whose boundaries can be expressed through natural-language definitions and examples.
It may be less suitable when the task depends on private expert knowledge absent from the gold examples, labels are highly ambiguous, or LLM API calls are prohibited.
At very large scale, token costs, provider limits, and retry overhead can also make direct LLM annotation inefficient.
In such settings, Goldsmith is better used to build a small, high-quality labeled set and a stable operational definition before training a supervised PLM or other task-specific model for large-scale labeling.

\section*{Ethical Considerations}

Goldsmith is intended to reduce the cost of building annotation workflows, not to replace expert judgment in high-stakes domains.
The workflow records model calls, prompt revisions, reviewer decisions, and export artifacts so that users can inspect and correct model behavior.
It sends task text, prompts, retrieved references, model predictions, and judge inputs to hosted LLM APIs; private, restricted, or sensitive corpora therefore require data-governance, provider-policy, and institutional review before use.
Our experiments use derived slices from existing OpenNER, CrossRE, and PHEE V2 artifacts, and users must follow the source datasets' licenses and redistribution terms.
Human review remains necessary because model predictions and judge scores are not authoritative labels, particularly in medical, legal, safety-critical, or other high-risk settings.

\section*{Acknowledgments}

This project was supported by funding from the 2026 Sun Yat-sen University Undergraduate Innovation and Entrepreneurship Training Program.
We thank Jie Qiu, Zhouxiaoyao Li, Beinuo Wang, and Xiaotong Liu for their substantial help during the early development of the project.

\bibliography{references}
\clearpage
\appendix

\section{Data Construction and Gold Selection}
\label{app:data}

The three task packages are constructed with the same principle: first build a deterministic task slice, then reserve a 15-item calibration set that maximizes coverage before matching the full-pool distribution.
The scripts write Rosetta/Prodigy-compatible JSONL records with source text, spans or relation labels, task identifiers, and metadata.

\subsection{OpenNER Typed Span Slice}
\label{app:openner-slice}
The OpenNER typed experiments use the local Rosetta-formatted OpenNER release after conversion from BIO files to JSONL.
The conversion procedure reconstructs token offsets and typed spans, then writes records with schema \texttt{rosetta.prodigy\_jsonl.v1}.
The deterministic multilingual slice is recorded with its seed, language coverage, and sampling statistics.

\begin{promptlisting}{OpenNER deterministic slice sampler}
Input: converted OpenNER JSONL files, seed, per_language = 50
For each language file:
  For each task:
    key = SHA256(seed, language, task_id)
  Sort tasks by (key, task_id)
  Select the first per_language tasks
Merge all selected language slices
Write all.jsonl and manifest.json
\end{promptlisting}

With seed 9 this yields 52 languages, 2,600 tasks, and 3,833 typed spans: 1,424 \(\mathrm{PER}\), 1,367 \(\mathrm{LOC}\), and 1,042 \(\mathrm{ORG}\).
Any binary \(\mathrm{NE}\) conversion is used only for OpenNER-derived PLM diagnostics, not as the official OpenNER task.

\subsection{OpenNER Gold Selection}
\label{app:openner-gold}
The OpenNER calibration set is selected by a deterministic coverage-first procedure.
Rows are featurized by language group, label-set state, length bin, and a stable SHA256 tie-breaker over \texttt{gold15}, \texttt{seed}, and \texttt{row\_id}.
Length bins use thresholds 40, 100, and 220 characters.

\begin{promptlisting}{Gold15 coverage-first selector}
Input: candidate rows R, target size n = 15
For each row r in R:
  group(r) = language or domain
  labels(r) = label set or relation label
  length_bin(r) = discretized text length
  tie(r) = SHA256(selector_name, seed, row_id)

Search feasible selected sets S of size n:
  Maximize group coverage, label coverage, and length-bin coverage
  Then minimize L1 distance between selected and full-pool distributions
  Then break ties by stable SHA256 order

Output: gold15.jsonl, remaining pool JSONL, selection report
\end{promptlisting}

The selection statistics record 2,600 candidate rows, 15 selected rows, 2,585 pool rows, 15 language groups covered, all four label states covered \((\mathrm{PER},\mathrm{ORG},\mathrm{LOC},\mathrm{NO\_LABEL})\), all four length bins covered, and no overlap between gold and pool files.

\subsection{CrossRE Pair-Level Slice}
\label{app:crossre-slice}
The CrossRE relation experiments use a compact package generated from the flattened sentence-level data.
Starting from flattened sentence-level CrossRE JSONL, the script groups rows by domain and computes a SHA256 key from \texttt{seed}, \texttt{domain}, and \texttt{doc\_key}.
For each of the six domains \(\{\)ai, literature, music, news, politics, science\(\}\), rows are sorted by \((\mathrm{key},\texttt{id})\) and the first 200 are selected.

\begin{promptlisting}{CrossRE pair-view construction}
Input: selected sentence-level CrossRE records
For each sentence record:
  Read entity mentions and gold relations
  Enumerate directed HEAD -> TAIL entity pairs
  If a gold relation exists for the directed pair:
    label = relation type
  Else:
    label = no_relation
  Write sentence, HEAD span/type, TAIL span/type, label
\end{promptlisting}

This produces 1,200 sentence-level tasks and a directed pair-level view with 39,764 entity-pair classification items.
The resulting manifest reports 35,275 \texttt{no\_relation} items, making positive-class metrics necessary alongside accuracy.

\subsection{CrossRE Gold Selection}
\label{app:crossre-gold}
CrossRE gold calibration sets are selected by a deterministic coverage-first procedure.
The main relation-classification experiments use the pair-level \texttt{rc\_pairs\_gold15} output.
For each row, the stable tie-breaker is a SHA256 hash over \texttt{crossre-gold15}, \texttt{seed}, \texttt{domain}, \texttt{identity}, and \texttt{row\_id}.
The selector first greedily builds a 15-item set that maximizes domain, relation-label, and length-bin coverage, then performs exhaustive single-swap local search until no swap improves the lexicographic objective.
The selection statistics record 39,764 candidates, 15 selected items, all six domains covered, 15 relation labels covered, all four length bins covered, and 39,749 remaining pool items.

\subsection{PHEE V2 Event-Argument Setting}
\label{app:phee}
PHEE V2 is used as a controlled event-argument extraction setting associated with pharmacovigilance event extraction \citep{phee2024chatgpt}.
Each item supplies a fixed event trigger anchor. The annotation interface requires argument spans for \texttt{Subject}, \texttt{Treatment}, and \texttt{Effect}, together with trigger-to-argument relations.
We report argument micro-F1, relation micro-F1, trigger pass rate, and joint exact match for initial zero-shot prompting, Goldsmith without review, and Goldsmith score-routed review.

\section{Evaluation Evidence Package}
\label{app:evidence}

\subsection{Model and Runtime Details}
\label{app:model-runtime}
Table~\ref{tab:model-runtime-details} lists the model families, run-facing model identifiers, access month, and decoding/runtime parameters used in the main LLM experiments.
Closed hosted models can change over time; for this reason we report the access month, provider-facing identifier, temperature, retry bound, concurrency, and candidate/search settings rather than treating a model nickname alone as a reproducible configuration.
Platform-call failures are retried up to the run bound, while successful calls with invalid structured output enter format repair before the item is requeued.

\begin{table*}[t]
  \centering
  \scriptsize
  \setlength{\tabcolsep}{3pt}
  \begin{tabular}{@{}p{0.16\textwidth}p{0.13\textwidth}p{0.16\textwidth}p{0.08\textwidth}p{0.08\textwidth}p{0.29\textwidth}@{}}
    \toprule
    Use & Provider family & Run-facing identifier & Access & Temp. & Main runtime settings \\
    \midrule
    Default annotation / judge & OpenAI & \texttt{gpt-5.5} & May 2026 & 0.0 / 0.0 & max tokens 4096; retry bound 20; concurrency 30 \\
    OpenNER model ablation & OpenAI & \texttt{gpt-5.4} & May 2026 & 0.0 / 0.0 & same score-routed review protocol as default \\
    OpenNER model ablation & Moonshot AI & \texttt{kimi-k2.5} & May 2026 & 0.0 / 0.0 & same first-pass protocol; pass1 completed-item probe \\
    OpenNER model ablation & DeepSeek & \texttt{deepseek-v4-pro} & May 2026 & 0.0 / 0.0 & same score-routed review protocol as default \\
    Prompt optimization & OpenAI & \texttt{gpt-5.5} & May 2026 & 0.3 / 0.0 & 30 rounds; 5 candidates; patience 5 \\
    \bottomrule
  \end{tabular}
  \caption{LLM backends and runtime configurations used for annotation, judging, model ablations, and prompt optimization.}
  \label{tab:model-runtime-details}
\end{table*}

\subsection{OpenNER Split Accounting}
\label{app:openner-split-accounting}
The main OpenNER score-routed review run contains 2,600 target items, 15 calibration items, and 193 reviewed feedback items.
After pass1, the lowest judge-score items are corrected as human review feedback and promoted to reviewed references.
The system then reruns annotation over the target corpus; reported score-routed review metrics are computed on the same held-out subset used by the without review and baseline rows after excluding calibration and reviewed feedback items.

\subsection{CrossRE Scope Accounting}
\label{app:crossre-scope-accounting}
All CrossRE rows in Table~\ref{tab:plm-budget-curve} are reported on the same 29,171-pair evaluation slice.
The Goldsmith workflow uses reviewed feedback and retrieved references during annotation, while final metrics are computed on this full pair-level evaluation slice.
The PLM references use a low-budget supervised setting: 200 training sentences are expanded into 10,593 training pairs before evaluation on the same 29,171-pair slice.
This appendix records how the pair-level slice, reviewed feedback, and PLM low-budget training split are constructed.

\subsection{GT-only RAG Reference Count Sweep}
\label{app:gt-rag-sweep}
This diagnostic OpenNER sweep isolates the number of retrieved similar references.
It uses the same optimized operational definition as the main OpenNER experiment, a fixed 500-example evaluation sample, and oracle gold-truth examples as the retrieval pool.
It retrieves only similar examples, with no reviewed model predictions included as references.
Performance improves substantially from \(k=0\) to moderate \(k\), peaks at \(k=14\) with \(0.8026\) micro-F1, and then declines slightly for larger \(k\).
This supports tuning the number of in-context references rather than assuming that more retrieved examples are always better.

\begin{table}[t]
  \centering
  \footnotesize
  \setlength{\tabcolsep}{5pt}
  \begin{tabular}{@{}rrrrrr@{}}
    \toprule
    \(k\) & Micro P & Micro R & Micro F1 & Macro F1 & Exact \\
    \midrule
    0 & 0.7173 & 0.7296 & 0.7233 & 0.8459 & 0.7360 \\
    1 & 0.7542 & 0.7652 & 0.7597 & 0.8611 & 0.7480 \\
    2 & 0.7569 & 0.7599 & 0.7584 & 0.8627 & 0.7600 \\
    3 & 0.7644 & 0.7704 & 0.7674 & 0.8660 & 0.7620 \\
    4 & 0.7894 & 0.7863 & 0.7878 & 0.8709 & 0.7720 \\
    5 & 0.7971 & 0.7876 & 0.7923 & 0.8752 & 0.7720 \\
    6 & 0.7849 & 0.7797 & 0.7823 & 0.8716 & 0.7700 \\
    7 & 0.7742 & 0.7823 & 0.7782 & 0.8690 & 0.7680 \\
    8 & 0.7807 & 0.7797 & 0.7802 & 0.8734 & 0.7760 \\
    9 & 0.7835 & 0.7876 & 0.7855 & 0.8725 & 0.7720 \\
    10 & 0.8048 & 0.7942 & 0.7995 & 0.8789 & 0.7780 \\
    11 & 0.7868 & 0.7889 & 0.7879 & 0.8767 & 0.7660 \\
    12 & 0.7836 & 0.7929 & 0.7882 & 0.8700 & 0.7640 \\
    13 & 0.7887 & 0.7929 & 0.7908 & 0.8713 & 0.7720 \\
    14 & \textbf{0.8059} & \textbf{0.7995} & \textbf{0.8026} & \textbf{0.8820} & \textbf{0.7860} \\
    15 & 0.7763 & 0.7691 & 0.7727 & 0.8778 & 0.7740 \\
    16 & 0.7997 & 0.7955 & 0.7976 & 0.8779 & 0.7760 \\
    17 & 0.7866 & 0.7929 & 0.7898 & 0.8733 & 0.7680 \\
    18 & 0.7925 & 0.7810 & 0.7867 & 0.8764 & 0.7740 \\
    19 & 0.7886 & 0.7876 & 0.7881 & 0.8701 & 0.7720 \\
    20 & 0.7857 & 0.7836 & 0.7847 & 0.8754 & 0.7700 \\
    \bottomrule
  \end{tabular}
  \caption{OpenNER GT-only diagnostic sweep over the number of retrieved gold references $k$.}
  \label{tab:gt-rag-k-sweep}
\end{table}

\subsection{Prompt-Optimization Trace}
\label{app:prompt-optimization-trace}
Table~\ref{tab:prompt-optimization-round-trace} reports the OpenNER prompt-optimization trace used for the main optimizer comparison.
All five optimizers use the same initial definition, 15 gold examples, candidate count \(5\), maximum \(30\) rounds, patience \(5\), and the same gold loss.
The table reports raw loss per round for rewrite-only, OPRO, and Goldsmith, plus the running best Goldsmith loss; the summary comparison also includes APE and PromptBreeder.
It also reports the approximate size of the accepted Goldsmith operational definition and the estimated tokens consumed by that Goldsmith optimization round.
The definition length is reported as an approximate token count using characters divided by four, while round tokens include candidate generation, gold evaluation, and repair calls.

\begin{table*}[t]
  \centering
  \scriptsize
  \setlength{\tabcolsep}{3pt}
  \begin{tabular*}{\textwidth}{@{\extracolsep{\fill}}rrrrrrr@{}}
    \toprule
    Round & Rewrite loss & OPRO loss & Goldsmith loss & Goldsmith best & Def. tok. & Round tok. \\
    \midrule
    0 & 25.5152 & 25.5152 & 25.5152 & 25.5152 & 433 & 0 \\
    1 & 24.5303 & 19.5303 & 16.9697 & 16.9697 & 450 & 95,848 \\
    2 & 22.3636 & 23.0152 & 12.5152 & 12.5152 & 521 & 98,151 \\
    3 & 23.1667 & 17.1970 & 12.1364 & 12.1364 & 578 & 104,963 \\
    4 & 23.1212 & 20.6667 & 12.1364 & 12.1364 & 671 & 112,011 \\
    5 & 25.2727 & 19.1515 & 9.5758 & 9.5758 & 740 & 120,960 \\
    6 & 23.1667 & 19.5303 & 7.4924 & 7.4924 & 768 & 125,760 \\
    7 & 22.0303 & 17.3485 & 5.2121 & 5.2121 & 822 & 127,690 \\
    8 & 23.6970 & 17.5303 & 4.2121 & 4.2121 & 907 & 136,372 \\
    9 & 24.6970 & 16.9242 & 3.8333 & 3.8333 & 930 & 140,810 \\
    10 & 23.1667 & 19.5303 & 4.2121 & 3.8333 & 943 & 143,974 \\
    11 & 22.0303 & 18.9242 & 4.2121 & 3.8333 & 1,010 & 146,753 \\
    12 & 24.3636 & 21.1212 & 3.8333 & 3.8333 & 1,101 & 154,491 \\
    13 & 26.9848 & 19.3788 & 4.2121 & 3.8333 & 1,101 & 162,326 \\
    14 & 23.8636 & 18.9242 & 4.2121 & 3.8333 & 1,101 & 160,502 \\
    15 & 27.8636 & 18.0606 & 4.2121 & 3.8333 & 1,151 & 157,544 \\
    16 & 28.5556 & 14.5909 & 2.1818 & 2.1818 & 1,208 & 169,331 \\
    17 & 28.5556 & 19.5303 & 2.1818 & 2.1818 & 1,208 & 167,254 \\
    18 & 29.3636 & 19.0152 & 1.8030 & 1.8030 & 1,252 & 171,998 \\
    19 & 27.4192 & 19.5303 & 2.1818 & 1.8030 & 1,252 & 178,037 \\
    20 & 28.5556 & 20.6667 & 1.8030 & 1.8030 & 1,168 & 176,575 \\
    21 & 27.4192 & 16.2121 & 3.3939 & 1.8030 & 1,200 & 168,229 \\
    22 & 27.4192 & 19.3485 & 1.8030 & 1.8030 & 1,200 & 177,896 \\
    23 & 28.5556 & 18.9242 & 2.1818 & 1.8030 & 1,225 & 173,501 \\
    24 & 28.1768 & 17.3485 & 1.8030 & 1.8030 & 1,309 & 177,479 \\
    25 & 27.4192 & 17.7273 & 1.8030 & 1.8030 & 1,372 & 185,558 \\
    26 & 28.5556 & 17.8788 & 1.8030 & 1.8030 & 1,384 & 187,319 \\
    27 & 28.5556 & 19.5303 & 1.8030 & 1.8030 & 1,384 & 193,772 \\
    28 & 28.5303 & 16.5455 & 1.8030 & 1.8030 & 1,384 & 192,780 \\
    29 & 29.3636 & 19.1515 & 2.1818 & 1.8030 & 1,375 & 189,262 \\
    30 & 29.6667 & 17.7273 & 2.1818 & 1.8030 & 1,444 & 194,873 \\
    \bottomrule
  \end{tabular*}
  \caption{Per-round OpenNER gold loss and token usage for LLM rewrite-only, OPRO, and Goldsmith. Definition tokens estimate the accepted Goldsmith definition length; round tokens estimate optimization and evaluation usage.}
  \label{tab:prompt-optimization-round-trace}
\end{table*}

\subsection{Gold-Loss Parameter Configuration}
\label{app:gold-loss-parameters}
The main text presents the loss symbolically so that the objective is not tied to arbitrary numeric scales. The implementation instantiates the following fixed configuration for the reported experiments. These values are held constant when comparing candidate definitions; they are not tuned separately for individual candidates or rounds.

\begin{table}[t]
  \centering
  \footnotesize
  \setlength{\tabcolsep}{4pt}
  \begin{tabular}{@{}llp{0.48\columnwidth}@{}}
    \toprule
    Parameter & Value & Role \\
    \midrule
    \(\rho\) & 100 & Reporting scale applied to the normalized gold loss \\
    \(\alpha_{\mathrm{miss}}\) & 1.2 & Weight on missed boundaries/relations \\
    \(\alpha_{\mathrm{extra}}\) & 1.0 & Weight on extra boundaries/relations \\
    \(\lambda_B\) & 0.75 & Boundary component of span loss \\
    \(\lambda_T\) & 0.25 & Type component of span loss \\
    \(\lambda_S\) & 0.5 & Span component of span--relation loss \\
    \(\lambda_R\) & 0.5 & Relation component of span--relation loss \\
    \(\lambda_{\mathrm{exact}}\) & 0.10 & Typed/span-structure exact-mismatch penalty \\
    \(\lambda_{\mathrm{stab}}\) & 0.05 & Inconsistency penalty for unstable repeated outputs \\
    \(\lambda_{\mathrm{trigger}}\) & 1.0 & Fixed-trigger contract violation penalty \\
    \(\lambda_{\mathrm{parse}}\) & 1.25 & Penalty for an unparseable model response \\
    \bottomrule
  \end{tabular}
  \caption{Gold-loss parameters used by the Rosetta implementation. The denominator safeguards \(\max(1,n)\) prevent division by zero for empty gold or prediction sets. For relation-only classification, label disagreement is the task loss and is not counted again as a separate exact-mismatch penalty.}
  \label{tab:gold-loss-parameters}
\end{table}

The boundary and relation rate weights are normalized by their sum, so their absolute scale does not affect the corresponding rate. Likewise, \(\lambda_B+\lambda_T=1\) and \(\lambda_S+\lambda_R=1\) make the component weights interpretable as relative contributions. The factor \(\rho\) is a monotonic reporting transformation: removing it would change only the displayed loss values, not candidate ordering or the acceptance decision. The implementation currently does not add a prompt-length penalty to the loss.

\subsection{CrossRE Prompt-Optimization Trace}
\label{app:crossre-prompt-trace}
The CrossRE pair-level optimizer starts with 8/15 calibration items passing and reaches 15/15 with zero gold loss at round 8.
The complete trace is shown in Table~\ref{tab:crossre-prompt-optimization-trace}; the run evaluated 40 candidate prompts under the same gold-loss protocol used for OpenNER.

\begin{table}[t]
  \centering
  \footnotesize
  \setlength{\tabcolsep}{6pt}
  \begin{tabular}{@{}lrr@{}}
    \toprule
    Round & Gold loss $\downarrow$ & Pass \\
    \midrule
    Initial & 46.6667 & 8/15 \\
    1 & 46.6667 & 8/15 \\
    2 & 46.6667 & 8/15 \\
    3 & 46.6667 & 8/15 \\
    4 & 33.3333 & 10/15 \\
    5 & 26.6667 & 11/15 \\
    6 & 13.3333 & 13/15 \\
    7 & 6.6667 & 14/15 \\
    8 & \textbf{0.0000} & \textbf{15/15} \\
    \bottomrule
  \end{tabular}
  \caption{CrossRE Goldsmith training trace on the 15-example gold calibration set, showing gold loss and pass count by optimization round.}
  \label{tab:crossre-prompt-optimization-trace}
\end{table}

\section{Baseline Implementation Details}
\label{app:baselines}

\subsection{Task-Stage Baselines}
\label{app:task-baselines}
The OpenNER LLM baselines use the same typed span interface as Goldsmith.
\textbf{Zero-shot LLM} receives the task schema and initial task description without gold examples, retrieval, review, or prompt training.
\textbf{Static ICL LLM} receives the same 15 gold examples as in-context references but does not optimize the operational definition and does not use score-routed review.
\textbf{Goldsmith without review} uses the optimized definition but reports the first annotation pass before human correction.
\textbf{Goldsmith score-routed review} uses the optimized definition, corrects low judge-score items as reviewed feedback, promotes them to reviewed references, and reruns annotation over the target corpus before reporting the second-pass result.

The OpenNER PLM baselines are low-budget token-classification models.
The training split contains 208 deterministic multilingual examples from the same typed OpenNER setting.
mBERT and XLM-R are fine-tuned as token classifiers over the typed \(\mathrm{PER}/\mathrm{ORG}/\mathrm{LOC}\) label space and evaluated by reconstructed typed spans.
These rows are intended as low-label supervised anchors, not as fully tuned large-data NER systems.

The CrossRE LLM setting is pair-level relation classification: the input supplies a sentence, a marked HEAD entity, a marked TAIL entity, entity types, and the candidate relation-label set.
The model predicts one relation label or \texttt{no\_relation}.
The typed-marker PLM baselines use the same given-pair interface.
The reported PLM setting selects 200 training sentences, expands them into 10,593 training pairs, and evaluates on 29,171 test pairs from the remaining 1,000 sentences.
BERT is the CrossRE anchor encoder, while RoBERTa-large and DeBERTa-v3 are stronger encoder variants under the same typed-marker input format.

\subsection{Prompt-Optimization Baselines}
\label{app:prompt-baselines}
The prompt-optimization comparison uses five optimizers under a shared search protocol: same initial OpenNER operational definition, same 15 gold examples, same candidate count or matched search budget, same loss, same acceptance rule, and the same stop policy.
\textbf{LLM rewrite-only} receives only the current operational definition and asks the LLM to produce candidate rewrites; it receives no scores, history, gold examples, failed examples, or textual gradients.
\textbf{OPRO} receives previous operational definitions and their loss-derived scores, but does not see source text, gold answers, model answers, or error types.
\textbf{APE} uses its candidate-generation and scoring implementation under the shared GPT-5.5 evaluation protocol.
\textbf{PromptBreeder} uses a budget-matched reproduction of the evolutionary search procedure.
\textbf{Goldsmith} summarizes high-loss gold errors into textual-gradient edit directions, asks an LLM editor for abstract candidate definitions, and accepts a candidate only when gold loss decreases.
The exact optimizer prompt templates and the final operational definitions are listed in Appendix~\ref{app:prompts}, so this section records only the implementation differences.

\subsection{Agent-Assisted Code Work}
\label{app:agent-code-work}
We used Codex with GPT-5.5 as an assistant for part of the code construction work.
The authors reviewed the resulting code and remain responsible for the final submission.

\section{Prompt Artifacts}
\label{app:prompts}
\noindent This appendix records the optimizer templates and operational definitions used in the experiments.
Fixed schema, parser contracts, and JSON output protocols remain part of the harness rather than the optimizable definition.
Concrete failed-example blocks are represented by placeholders because they are training-only feedback, not deployed prompt text.

\begin{promptlisting}{Optimizer implementation: LLM rewrite-only baseline}
System prompt:
You are a baseline prompt editor. Return only candidate concept definitions in the requested tags.

User prompt template:
Rewrite the concept prompt below into {candidate_count} better and more precise candidate prompts.

Current optimizable definition:
{current_operational_definition}

Rules:
1. Use only the current prompt above. Do not ask for or invent examples, scores, failure details, history, gradients, logs, or diagnostics.
2. Each candidate must be a standalone concept definition usable for annotation.
3. Include only task definition and semantic boundary guidance.
4. Do not include label sets, JSON schema, annotation markup, parser rules, output format, repair instructions, examples, explanations, or revision notes.
5. If two candidates are similar, make the later one shorter or more precise.

Return exactly this tagged format:
<CANDIDATE_1>...</CANDIDATE_1>
...
<CANDIDATE_k>...</CANDIDATE_k>
\end{promptlisting}

\begin{promptlisting}{Optimizer implementation: OPRO}
System prompt:
You are an OPRO prompt optimization assistant. Return exactly one new concept definition inside <INS>...</INS>.

User prompt template:
You are using Optimization by PROmpting (OPRO) to improve an annotation concept prompt.

Below are historical concept prompts and their scores on the same 15 gold examples. Higher score is better.

{prompt_score_history_only}

Current optimizable concept prompt:
{current_operational_definition}

Generate one different concept prompt that may obtain a higher score.

Rules:
1. Use only the prompt-score history above; do not ask for or invent error details.
2. Keep the new prompt concise. If two prompts are likely tied, prefer the shorter one.
3. Include only task definition and semantic boundary guidance.
4. Do not include label sets, JSON schema, annotation markup, parser rules, output format, repair instructions, examples, logs, or explanations.
5. Return only:
<INS>
new concept prompt
</INS>
\end{promptlisting}

\begin{promptlisting}{Optimizer implementation: Goldsmith textual-gradient editor}
Stage 1 system prompt:
You are the Ours textual-gradient critic. Return concise textual gradients only, not a final prompt.

Stage 1 user prompt template:
Generate textual gradients for improving the current annotation concept prompt.

training_feedback_only=true

Current optimizable prompt:
{current_operational_definition}

Current loss:
{current_gold_loss}

Overall failure summary:
{failure_summary}

Failed gold examples with source text, gold answer, model answer, and error type:
{top_high_loss_gold_errors}

Return 3-5 concise textual gradients. Each gradient should explain why the current prompt caused the observed errors and what semantic boundary should change.

Output requirements:
1. Output only numbered or bulleted textual gradients.
2. Do not generate the final prompt.
3. Do not copy source text, gold answers, model answers, labels, JSON schema, annotation markup, parser rules, output format, or repair instructions into a reusable prompt.

Stage 2 system prompt:
You are the Ours prompt editor. Return only the final usable concept definition.

Stage 2 user prompt template:
Rewrite the concept prompt by applying one Ours textual gradient.

training_feedback_only=true

Current optimizable prompt:
{current_operational_definition}

Textual gradient {gradient_index}:
{textual_gradient}

Variant index: {variant_index}

Output requirements:
1. Return only the concept definition body usable in the next annotation round.
2. Apply the gradient as an abstract semantic boundary change.
3. Keep the prompt concise; if two rewrites are equivalent, prefer the shorter one.
4. Include only task definition and semantic boundary guidance.
5. Do not output labels, JSON schema, annotation markup, parser rules, output format, repair instructions, examples, explanations, failed example ids, or revision logs.
6. Do not copy source text, gold answers, or model answers.
\end{promptlisting}

\begin{promptlisting}{OpenNER final definition: LLM rewrite-only baseline}
Annotate every explicit proper name that names a single person, organization, or location. The name may appear in any language or script and must be assigned one, and only one, semantic type.

Include as person names all proper names for individual people: full names, one-token names, given names, surnames, initials, short forms, aliases, nicknames, and established personal-name compounds. Do not include titles, honorifics, ranks, offices, jobs, kinship terms, or descriptive phrases unless they are fixed within the written name.

Include as organization names proper names of collective or institutional entities, such as companies, institutions, agencies, government bodies, political parties, associations, banks, schools, religious bodies, committees, and named programs, initiatives, campaigns, or acronyms denoting those entities. Do not include generic descriptions, unnamed subunits, document or publication titles, products, software, models, versions, technical abbreviations, or formal phrases that are not entity names.

Include as location names proper names of geographic or geopolitical places, such as countries, cities, towns, regions, states, provinces, counties, districts, neighborhoods, territories, conventional place names, and named political areas. Do not include common place nouns, vague or relative locations, coordinates, dates, numbers, or descriptive geography unless it is part of an established name.

Use the exact contiguous text span only. Preserve all original characters and formatting features, including spelling, script, case, diacritics, punctuation, spaces, hyphens, inflectional endings, possessives, particles, articles, connectors, and acronyms. Do not normalize, translate, expand, shorten, correct, or infer. Annotate complete established multiword names as single spans, but split adjacent coordinated or listed names unless the full expression is a conventional name. In CJK or other unspaced scripts, mark each complete name separately. If none appear, annotate nothing.
\end{promptlisting}

\begin{promptlisting}{OpenNER final definition: OPRO}
Multilingual NER for explicit surface proper names of persons, organizations, and places.

Annotate only contiguous text that is literally present and functions as the name of a particular entity. Do not annotate common descriptions, roles, offices, titles, occupations, kinship terms, categories, adjectives, nationalities, demonyms, dates, quantities, events, laws, reports, documents, products, releases, standards, slogans, technical terms, addresses, or other non-name phrases unless they are fixed within an eligible name.

PER: individual human names, real or fictional, including full names, given names, surnames, initials, aliases, nicknames, and shortened later mentions. Exclude honorifics, role words, professions, offices, kinship words, and descriptive references outside the name.

ORG: proper names of organized bodies or organization-like entities, including companies, institutions, government bodies, agencies, ministries, political parties, banks, schools, universities, teams, media outlets, nonprofits, military units, and formally named programs, projects, initiatives, or campaigns. Acronyms are ORG only when used as organization names. Exclude generic departments, administrative descriptions, document/law/report titles, products, releases, events, and abbreviations that do not name an organization.

LOC: proper names of geographic or geopolitical places, including countries, cities, towns, regions, states, provinces, districts, territories, landmarks, facilities used as places, natural features, seas, rivers, mountains, bodies of water, and conventional named areas. Bare place names are LOC even when used metonymically; a complete written name of a government, institution, company, team, or other organized body is ORG.

Choose the minimal complete contiguous name exactly as written. Include name-internal articles, particles, prepositions, conjunctions, initials, hyphens, apostrophes, clitics, possessives, inflection, case endings, and script-specific name material; exclude surrounding punctuation, determiners, external prepositions, modifiers, appositives, explanations, titles, and role words. Keep conventional multiword names together, do not mark substrings inside longer eligible names, split coordinated or listed names into separate spans, and never translate, normalize, romanize, correct, expand, or infer.
\end{promptlisting}

\begin{promptlisting}{OpenNER final definition: Goldsmith}
Multilingual typed named-entity span detection for explicit mentions of persons, organizations, and locations.

Mark only explicit named-entity mentions in the source text and assign the single applicable type. Use dataset-style minimalism: if a name-like span is ambiguous between a person, title component, venue descriptor, facility name, event/work title, or generic description, exclude it unless the context clearly establishes a target entity. Keep only spans with clear entity evidence.

Person mentions include explicit personal names, surnames, shortened personal references, multi-token personal names, and explicit named or vocative references to a deity or sacred personal figure when used as a personal addressee or name. Include transliterated, romanized, dialectal, or informal-script forms meaning God, Lord, or a named sacred figure only when they function as direct sacred personal references. For sacred references, mark only the core name or direct reference; exclude epithets, praise formulas, descriptive complements, honorific expansions, and devotional descriptions unless fixed internal parts of the name. Exclude titles, occupations, honorifics, descriptive phrases, generic human or divine descriptions, and capitalized/name-like strings that merely resemble names. Suppress person-shaped names inside creative-work titles, media listings, slogans, headings, releases, screenings, or other non-entity phrases unless the syntax clearly refers to an actual human or sacred personal participant. Full personal names in ownership, attribution, sale, purchase, transfer, or agentive clauses remain person mentions when clearly referring to people.

Organization mentions include named institutional actors and standing organizational bodies such as institutions, agencies, companies, banks, departments, committees, acronyms for such bodies, and conventional sovereign state bodies such as armed forces, militaries, and police. Otherwise require a distinctive proper-name component, jurisdictional name, or unambiguous standing-body identity. Exclude generic institutional nouns and descriptions, including phrases meaning government office, bureau, department, education bureau, or similar institutional types without distinctive naming or clear standing-body reference. Exclude numbered legislative terms, policies, methods, courses, releases, documents, events, festivals, products, training packages, schemes, initiatives, hubs, market products, and price or index acronyms unless they clearly name an organization. Do not label a facility, association, venue, harbor installation, landmark, or institutional-looking name as an organization when the context uses it as a physical place, vicinity, embarkation point, street/harbor reference, or topographic landmark.

Location mentions include clear geographic and geopolitical names, including countries, regions, cities, states or provinces, conventional multi-word place names, named geopolitical areas, and named physical venues or harbor/landmark places when used locatively or topographically. Prefer the bare proper place name inside broader descriptive, ideological, political, cultural, adjectival, or locative phrases. Do not include following generic place nouns or common/case-marked locative words such as square, street, plaza, district, area, office, department, or similar unless the full phrase is a conventional official place name. Be conservative with short single-token place-like strings before or modifying generic venue, facility, event, program, or title nouns: mark them only when they are clearly independent geographic or topographic place names, not facility-name components, event titles, descriptive modifiers, or proper-looking descriptors. Exclude generic location nouns, dates, quantities, adjectival demonyms, descriptive place words, and generic government or institutional common nouns.

Use exact surface substrings from the source text. Preserve spelling, accents, punctuation, script, spacing, and inflection. Keep conventional contiguous multi-token names as one span, including internal particles, articles, connectors, hyphens, acronyms, and script-specific case or possessive material. Split adjacent coordinated names into separate spans. For spaced scripts or tokenized text containing multiple place names, prefer separate complete place-name spans unless an internal connector forms one conventional name.

Do not translate, normalize, romanize, correct spelling, infer absent entities, or expand a later short mention beyond its actual surface form. If no explicit, clearly supported mention of the target entity types appears, return no spans.
\end{promptlisting}

\begin{promptlisting}{OpenNER initial task prompt}
Multilingual OpenNER typed named-entity span detection.

Mark explicit person, organization, and location names in the source text.
Assign each span exactly one label: PER, ORG, or LOC.

Use exact contiguous surface spans only.
If no target entity is explicitly mentioned, return no spans.
\end{promptlisting}

\begin{promptlisting}{OpenNER deployed final task prompt}
Multilingual OpenNER typed named-entity span detection.

Mark every explicit named-entity mention in the source text and assign exactly one label: PER, ORG, or LOC.

PER includes personal names, surnames, shortened person mentions, and multi-token person names. Exclude titles, occupations, honorifics, and generic human descriptions unless they are part of the name.

ORG includes named institutional actors and organizational bodies such as organizations, institutions, agencies, companies, banks, departments, committees, and their acronyms. Exclude policies, methods, courses, releases, documents, events, festivals, products, training packages, and official-sounding schemes or initiatives unless they clearly name an organization or standing organizational body.

LOC includes geographic and geopolitical names: countries, regions, cities, states or provinces, conventional multi-word place names, and named geopolitical areas. Prefer the bare geopolitical name when it appears inside a broader descriptive, ideological, political, cultural, or adjectival country phrase. Do not include such modifiers in the LOC span unless the full phrase is a conventional official place name. Exclude generic location nouns, dates, quantities, demonyms used only adjectivally, and descriptive place words unless they are inseparable parts of the proper name.

Use exact surface substrings from the source text. Preserve spelling, accents, punctuation, script, spacing, and inflection. Keep conventional contiguous multi-token names as one span, including internal particles, articles, connectors, hyphens, acronyms, and script-specific case or possessive material. Split adjacent coordinated names into separate spans. For CJK or tokenized text with spaces between location names, prefer separate complete place-name spans unless an internal connector forms one conventional name.

Do not translate, normalize, romanize, correct spelling, infer absent entities, or expand a later short mention beyond its actual surface form. If no explicit PER, ORG, or LOC mention appears, return no spans.
\end{promptlisting}

\begin{promptlisting}{CrossRE initial task prompt}
Cross-domain relation classification for English CrossRE entity pairs.

Classify only the provided directed HEAD -> TAIL entity pair.
The sentence, entity spans, entity types, and candidate relation labels are given.
Choose the relation label supported by local sentence evidence, or choose no_relation if no listed relation is expressed.
Do not discover entities, scan other pairs, reverse direction, or infer from background knowledge.
\end{promptlisting}

\begin{promptlisting}{CrossRE deployed final task prompt}
Cross-domain relation classification for English CrossRE directed entity pairs.

Classify only the single provided directed HEAD -> TAIL pair. The sentence and the HEAD/TAIL surface spans are fixed anchors. Do not discover entities, scan other pairs, reverse direction, add inverse edges, or infer from background knowledge.

Default decision rule:
- Return a positive relation only when the sentence gives clear local evidence that the directed HEAD entity has that relation to the TAIL entity.
- If the evidence is implicit, merely topical, based on broad world knowledge, based only on co-occurrence, or could apply equally to many nearby entities, choose no_relation.
- Prefer no_relation over a weak positive label. Precision is more important than guessing a plausible relation.
- Do not use `related-to` as a fallback for vague association. Use `related-to` only when the sentence explicitly states a contextual relation and no more specific label fits.

Label boundaries:
- `role`: explicit role, office, job, function, candidacy, lineup/team service, participation role, or acted/performed-as capacity.
- `physical`: explicit physical location, venue, residence, presence, geographic containment, or geopolitical containment.
- `general-affiliation`: explicit membership, association, political/organizational affiliation, genre identity, or group participation when no more specific role or structural relation is expressed.
- `part-of`: explicit structural or compositional inclusion: component, member, subgroup, subgenre, or item within a larger entity.
- `artifact`: creator, producer, author, developer, publisher, performer, or organization/person behind a specific work, product, book, album, song, algorithm, award, or artifact-like entity.
- `named`: alias, acronym, alternate name, renamed form, surname/full-name identity, or appositive naming.
- `temporal`: explicit relation to a date, period, competition edition, historical event, dated meeting, conference, or time-bearing context.
- `win-defeat`: winning, defeating, receiving an award, election outcome, or one side beating another.
- `origin`: founder, originator, source, inventor, derivation, originating entity, or granting/issuing source.
- `type-of`: explicit type/category relation.
- `usage`: explicitly uses, plays, employs, chooses, configures, or performs with an instrument, tool, technology, system, parameter, or similar used entity.
- `opposite`: explicit opposition, rivalry, conflict, battle, dispute, or adversarial relation.
- `topic`: a work, statement, degree, award, system, meeting, event, or document is explicitly about, in, concerning, or specialized in the TAIL topic/entity.
- `compare`: explicit comparison or contrast.
- `social`: explicit meeting, negotiation, interpersonal interaction, or direct person-to-person contact.
- `cause-effect`: one entity/event explicitly causes, results in, leads to, or produces another.

When multiple labels seem possible, choose the most specific label supported by explicit sentence evidence. If none is clearly supported for HEAD -> TAIL, output no_relation.
\end{promptlisting}

\end{document}